\documentclass[runningheads]{llncs}

\usepackage{eccv}

\usepackage{eccvabbrv}

\usepackage[T1]{fontenc}
\usepackage{lmodern}
\usepackage{graphicx}
\graphicspath{{figures/}{./}{../}}
\usepackage{booktabs}
\usepackage{multirow}
\usepackage[table]{xcolor}
\usepackage{amsmath,amssymb,bm}
\usepackage{placeins}
\usepackage{float}
\usepackage[accsupp]{axessibility}  % Improves PDF readability for those with disabilities.

\usepackage{hyperref}
\hypersetup{hidelinks}

\begin{document}

% ---------------------------------------------------------------
\title{CRISP: Calibration-Aware Visual State Space Duality for Remote Sensing Semantic Segmentation}
\titlerunning{CRISP for Remote Sensing Semantic Segmentation}

\author{Kangning Wang\inst{1,2} \and
Haopeng Zhang\inst{1,3,4}\thanks{Corresponding author.} \and
Zhiguo Jiang\inst{1}}
\authorrunning{K.~Wang et al.}

\institute{Tianmushan Laboratory, Beihang University, Hangzhou 311115, China \and
Hangzhou International Innovation Institute, Beihang University, Hangzhou 311115, China \and
School of Astronautics, Beihang University, Beijing 102206, China\\
\email{zhanghaopeng@buaa.edu.cn} \and
Key Laboratory of Spacecraft Design Optimization and Dynamic Simulation Technology,\\
Ministry of Education, Beijing 102206, China}

\maketitle

\begin{abstract}
State space models, especially Visual State Space Duality (VSSD), have emerged as efficient linear-time alternatives to Transformers for dense visual tasks. However, we observe that VSSD compresses spatial context into a global aggregation that suppresses high-frequency responses, causing excessive boundary smoothing in remote sensing semantic segmentation. To address this, we propose CRISP, a calibration framework with two components. Its core, the Duality Calibration Operator (DCO), restores local contrast and boundary responses through residual injection and frequency calibration within the VSSD backbone, without altering its linear complexity. To retain the recovered detail, an Orthogonal Multi-Prototype (OMP) head assigns multiple orthogonally constrained prototypes per class to model large intra-class variance. Extensive experiments on Potsdam, Vaihingen, and LoveDA show that, with only $\sim$30M parameters, CRISP achieves consistent gains in mean F1 (mF) and mIoU while remaining competitive with state-of-the-art methods. Code is available at \url{https://github.com/crazylifeha/crisp}.
\end{abstract}

\section{Introduction}
High-resolution remote sensing image semantic segmentation is a fundamental task in Earth observation, vital for urban planning, disaster monitoring, and environmental assessment~\cite{yuan2021review, cheng2017remote}. Unlike natural images, remote sensing imagery features exceedingly high spatial resolutions with dense, small-scale structures and complex category boundaries. Convolutional Neural Networks (CNNs) have been widely adopted for their local representation capability~\cite{fu2019dual, wang2022unetformer}, and Vision Transformers (ViTs) further advanced accuracy through long-range dependency modeling~\cite{dosovitskiy2020image, strudel2021segmenter}. However, the quadratic complexity of self-attention in sequence length~\cite{liu2021swin} causes severe memory and computation bottlenecks at the ultra-high resolutions typical of remote sensing.

\begin{figure*}[t]
  \centering
  \begin{minipage}{0.49\textwidth}
    \centering
    \includegraphics[width=\linewidth]{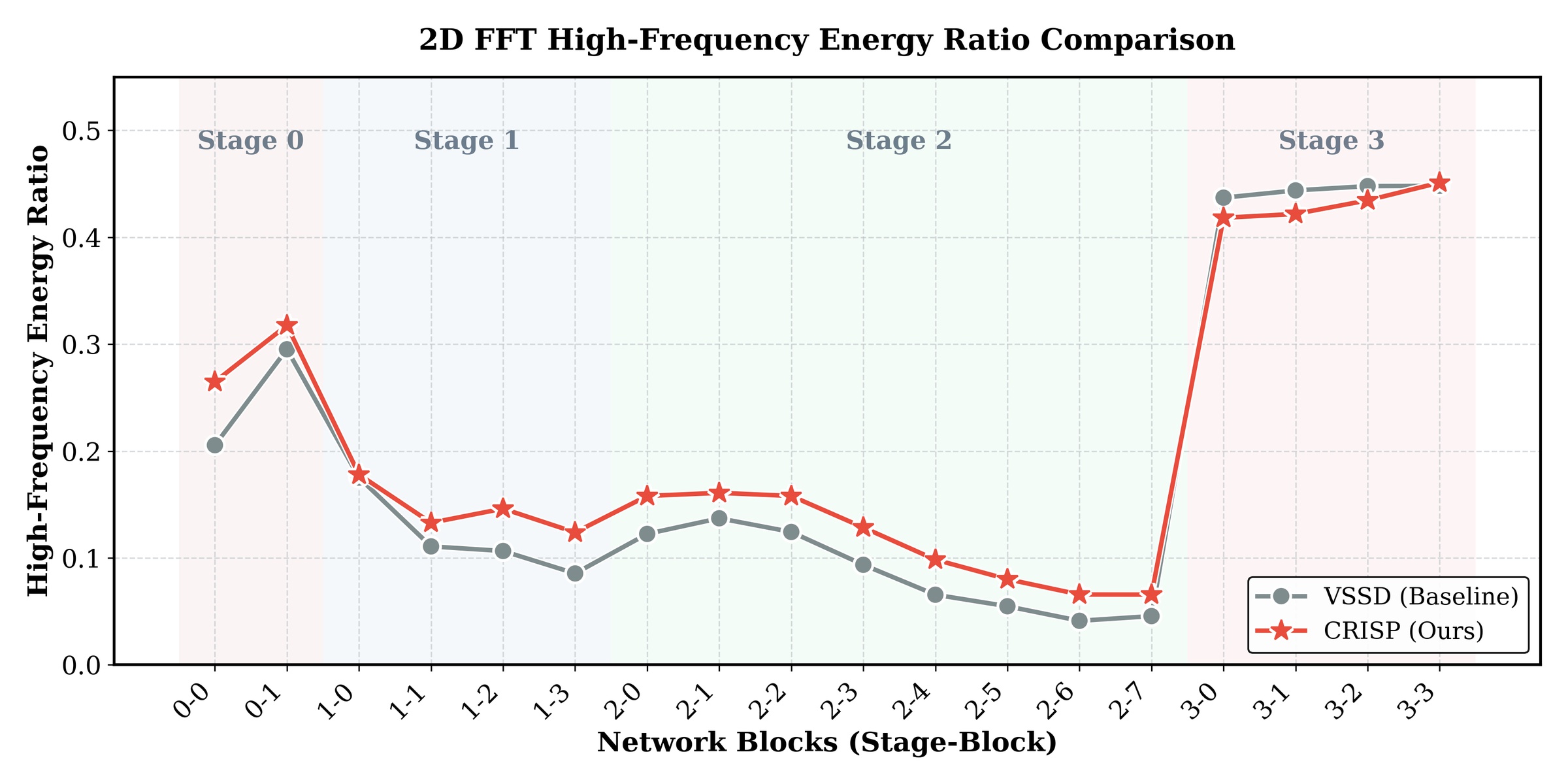}
    \textbf{(a)} High-frequency energy ratio.
  \end{minipage}\hfill
  \begin{minipage}{0.49\textwidth}
    \centering
    \includegraphics[width=\linewidth]{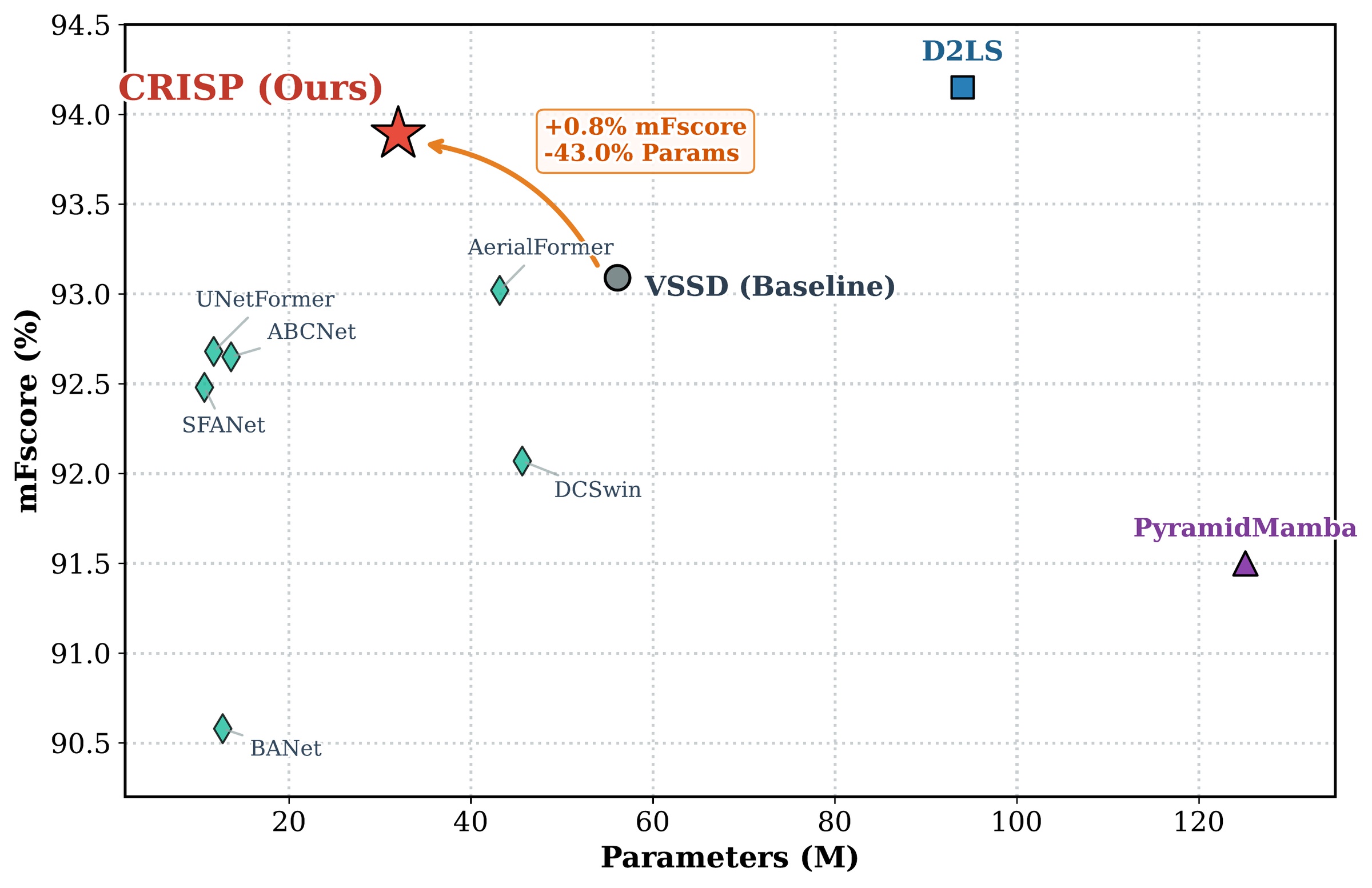}
    \textbf{(b)} Performance--complexity tradeoff on Potsdam.
  \end{minipage}
  \caption{Feature smoothing and efficiency analysis. (a) High-frequency energy ratio across backbone layers. (b) Performance--complexity tradeoff on Potsdam.}
  \label{fig:hf_and_tradeoff}
\end{figure*}

Recently, State Space Models (SSMs), particularly the selective state space-based Mamba~\cite{gu2024mamba} and its upgraded version Mamba-2, which introduced State Space Duality (SSD)~\cite{dao2024transformers}, have emerged as a promising alternative. SSD theoretically unifies sequence modeling and attention mechanisms; however, its causal scan assumes a single ordering of the input, which does not naturally fit 2D images. To alleviate this, recent studies proposed the Visual State Space Duality (VSSD) model~\cite{shi2025vssd}. Rather than a single causal pass, VSSD performs \emph{multi-directional scanning} and fuses the results into a global, order-agnostic aggregation, transferring the linear-complexity advantage of SSD to visual feature extraction. Nevertheless, our empirical studies reveal that this global aggregation tends to be unstable when characterizing boundaries and small-scale objects in remote sensing images enriched with complex high-frequency details. To quantify this phenomenon, we evaluate the baseline from two perspectives: (i) boundary-sensitive metrics (Boundary F-score), and (ii) feature spectral energy ratio ($E_{\text{high}}/E_{\text{low}}$, executing 2D FFT on intermediate backbone layers). As shown in Fig.~\ref{fig:hf_and_tradeoff}(a), the high-frequency energy ratio decays progressively across backbone layers, confirming the systematic suppression of high-frequency responses caused by the global normalized aggregation. Addressing this suppression, as we show later, lets CRISP reach a favorable performance--complexity trade-off (Fig.~\ref{fig:hf_and_tradeoff}(b)).

To address this without compromising VSSD's linear complexity, we propose CRISP, which introduces an aggregation-aligned calibration branch for targeted compensation rather than disrupting the normalized aggregation structure. CRISP couples a Duality Calibration Operator (DCO) in the backbone with an Orthogonal Multi-Prototype (OMP) head in the decoder, which jointly preserve intra-class multi-modal structures from pixel-level features to the final classification.

In summary, the main contributions of this paper are as follows:
\begin{itemize}
  \item We propose the Duality Calibration Operator (DCO), the first residual calibration for the VSSD aggregation, which reuses the backbone's aggregation weights to recover attenuated high-frequency responses while preserving linear complexity.
  \item We further introduce the Orthogonal Multi-Prototype (OMP) head, which learns orthogonally regularized sub-prototypes end-to-end to preserve the intra-class multi-modal structure recovered by DCO.
  \item We demonstrate that CRISP attains accuracy on par with state-of-the-art methods on Potsdam, Vaihingen, and LoveDA at only $\sim$30M parameters, achieving a markedly better accuracy--efficiency trade-off than prior methods.
\end{itemize}

\section{Related Work}
We organize prior work along four axes that together situate CRISP: (i) remote sensing semantic segmentation, (ii) visual state space models, (iii) frequency- and boundary-preserving modeling, and (iv) prototype-based decoders. The first two establish the backbone landscape CRISP builds on; the last two delineate the specific gap CRISP fills.

\subsection{Semantic Segmentation in Remote Sensing}
Remote sensing semantic segmentation assigns a category to every pixel of high-resolution overhead imagery, where dense small objects and intricate boundaries make the preservation of fine spatial detail essential.
Early CNN-based methods enlarge the effective receptive field while retaining locality: DANet~\cite{fu2019dual} couples spatial and channel attention to model long-range context, OCR~\cite{yuan2020object} aggregates object-contextual representations, and UNetFormer~\cite{wang2022unetformer} attaches a lightweight Transformer decoder to a CNN encoder for real-time urban-scene parsing.
With Vision Transformers, global dependency modeling improved accuracy---from plain ViT~\cite{dosovitskiy2020image} and SegFormer~\cite{xie2021segformer} to hierarchical Swin~\cite{liu2021swin} and mask-classification heads such as MaskFormer~\cite{cheng2021maskformer}---but at quadratic cost. To recover locality, hybrid CNN--Transformer designs~\cite{fan2023hybrid} interleave convolution and self-attention.
The shared limitation of these Transformer-based backbones is computational: the quadratic cost of self-attention in sequence length becomes memory-prohibitive at the ultra-high resolutions typical of remote sensing, which is precisely what motivates linear-complexity state-space alternatives.

\subsection{Visual State Space Models}
State Space Models (SSMs) originate in control theory: S4~\cite{gu2022s4} models long sequences with a structured linear-time recurrence, and Mamba~\cite{gu2024mamba} makes the recurrence input-dependent (selective) so it can route information content-adaptively. Mamba-2~\cite{dao2024transformers} establishes State Space Duality (SSD), showing this recurrence is equivalent to a masked linear-attention form; this connects SSMs to the broader linear-attention family~\cite{katharopoulos2020transformers, han2023flatten}.
Adapting SSMs to 2D images hinges on the \emph{scanning strategy} used to impose an order on pixels. Vision Mamba~\cite{zhu2024vision} applies a bidirectional scan, VMamba~\cite{liu2024vmamba} a four-directional cross-scan, PlainMamba~\cite{yang2024plainmamba} a continuous space-filling scan, and EfficientVMamba~\cite{pei2025efficientvmamba} an atrous selective scan for efficiency. Remote sensing backbones follow the same paradigm: RSMamba~\cite{chen2024rsmamba} and RS3Mamba~\cite{ma2024rs3mamba} tailor the scan paths of selective Mamba to overhead imagery, while CVMH-UNet~\cite{cao2025cvmh} even adds an explicit multi-frequency, multi-scale fusion module to counter detail loss---yet this remains an external add-on bolted onto an otherwise causal backbone. All of these variants still run \emph{causal} recurrences along each direction and approximate global context by summing several scans.
VSSD~\cite{shi2025vssd} departs from this line: instead of relying on a single per-direction causal ordering, it fuses multi-directional responses into a single global, position-agnostic aggregation, thereby reducing scan redundancy and roughly halving the scanning cost. We build on VSSD as our backbone for this reason.
Crucially, however, neither line examines the \emph{spectral} side effect of the global aggregation itself: the very step that yields linear cost acts as a low-pass operator, collapsing the directional scans into a near-uniform global summary and attenuating the high-frequency responses that encode boundaries and small objects. CRISP targets exactly this aggregation step, and is orthogonal to the choice of scan path or backbone topology.

\subsection{Frequency- and Boundary-Preserving Modeling}
To counter over-smoothing and recover lost high-frequency detail, prior methods can be grouped by \emph{where} they reinject the high-frequency signal.
(i) \emph{Spatial post-processing} sharpens the prediction after the backbone: PointRend~\cite{kirillov2020pointrend} adaptively re-samples uncertain boundary points to render crisper masks. Such methods correct the output mask but leave the already-smoothed backbone features untouched.
(ii) \emph{Frequency-domain operators} reinject high frequencies through a parallel spectral path: FreqMamba~\cite{zou2024freqmamba} couples Mamba with a frequency-domain branch for deraining, SpectralMamba~\cite{yao2024spectralmamba} performs spectral-domain scanning for hyperspectral data, and generic spectral modules such as Fast Fourier Convolution~\cite{chi2020fast} and Global Filter Networks~\cite{rao2021global} learn filters directly in the Fourier domain. Closer to dense prediction, FreqFusion~\cite{lin2024freqfusion} performs frequency-aware feature fusion on the decoder side to sharpen boundaries.
(iii) \emph{Locality restoration} reintroduces local detail in the spatial domain: LocalMamba~\cite{huang2024localmamba} adds windowed selective scans and parallel convolutions to recover the fine structure suppressed by global mixing.
(iv) \emph{Feature calibration} corrects or re-weights intermediate features instead of adding a spectral path: SFC~\cite{zhao2024sfc} shares feature calibration across classes for weakly-supervised segmentation, and CSFCal~\cite{li2023csfcal} jointly calibrates context and spatial features for real-time segmentation. CRISP belongs to this calibration family in spirit, but is the first to calibrate the SSM aggregation itself.
Effective as they are, these methods share two costs: they attach a \emph{separate} spectral branch, decoder-side fusion module, or convolutional path, adding parameters and FLOPs, and they break the hardware-friendly continuity of the linear SSM scan.
CRISP differs in both \emph{where} and \emph{what} it calibrates. Rather than appending a generic Fourier/DCT/wavelet module or a decoder-side fusion stage, our DCO calibrates the VSSD directional aggregation \emph{in place}: it reuses the aggregation weights already computed by the backbone to inject a state-level residual that compensates the discretization-induced aliasing, restoring high-frequency responses while preserving $\mathcal{O}(L)$ complexity and the native scan structure.

\subsection{Prototype-Based Decoders}
The decoder determines whether recovered fine detail survives to the final logits. Conventional heads such as DeepLabv3+~\cite{chen2018encoder} and UPerNet~\cite{xiao2018unified} are effectively \emph{single-prototype} classifiers: each class is represented by one weight vector, so a pixel is labeled by its similarity to that single center. This collapses intra-class variation onto one direction and is ill-suited to remote sensing, where the same category (\eg, buildings with different roof materials) exhibits very different spectral signatures.
Multi-center decoders relax this assumption. Center-guided classification~\cite{zhang2026center} maintains several class centers for remote sensing, NAPG~\cite{fan2025napg} groups neighborhood-assisted multi-prototypes, and prototypical contrastive methods~\cite{yu2025multiple} learn multiple prototypes per class. These approaches, however, typically obtain their centers via offline clustering or auxiliary contrastive training, which complicates end-to-end optimization and couples the decoder to a separate clustering stage.
Our Orthogonal Multi-Prototype (OMP) head instead learns $K$ class-wise sub-prototypes \emph{end-to-end}. Building on ideas from prototypical learning~\cite{snell2017prototypical}, orthogonal-prototype projection~\cite{liu2023pop}, and generalized few-shot remote sensing~\cite{li2024generalized}, OMP combines an intra-class orthogonality constraint with an inter-class angular margin and a mutual-information objective to keep the sub-prototypes diverse and non-redundant. This lets OMP preserve the fine-grained, multi-modal structure that DCO restores, without any offline clustering.

\section{Methodology}

\begin{figure}[ht]
  \centering
  \includegraphics[width=\linewidth]{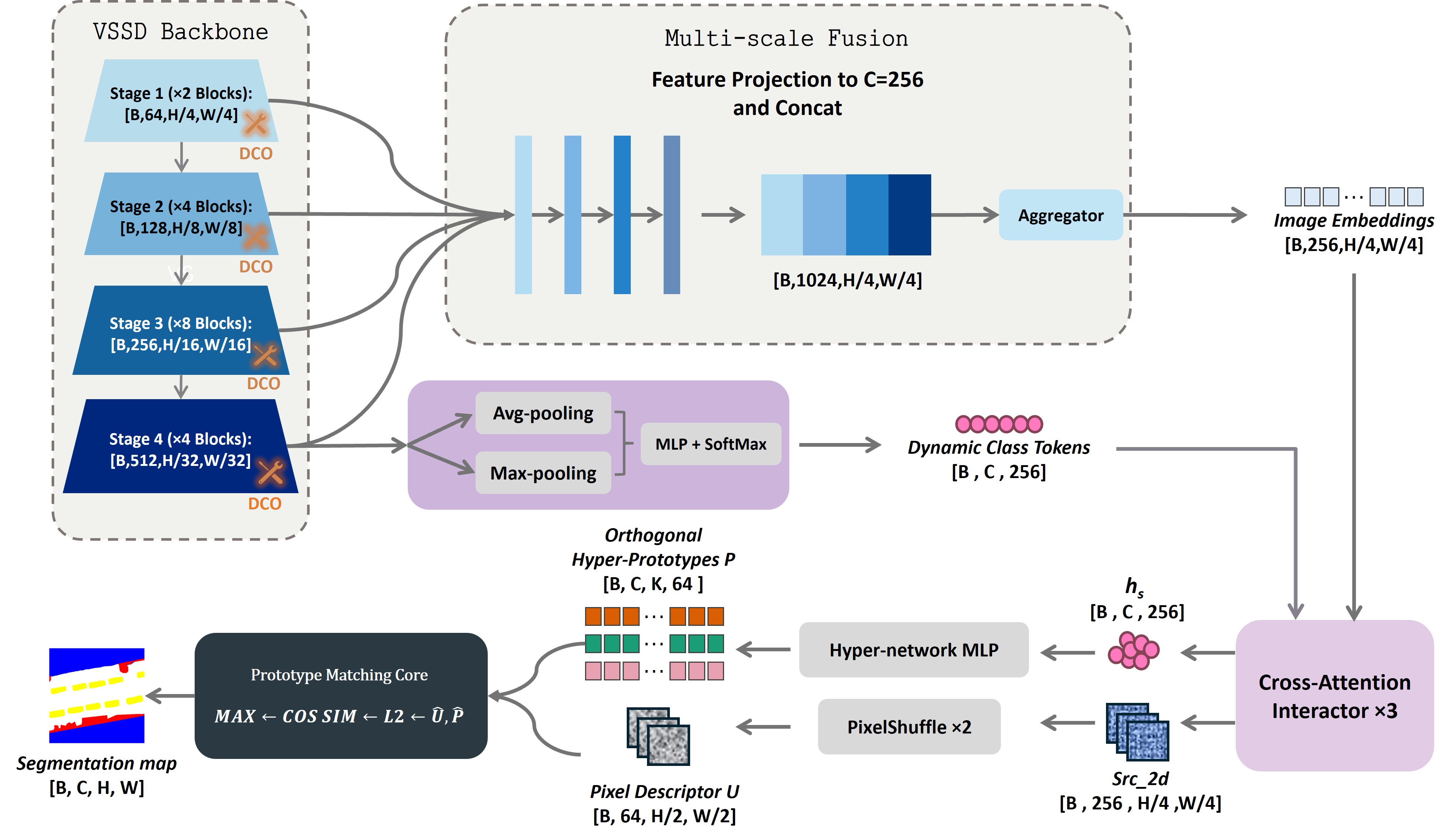}
  \caption{Overall CRISP architecture. DCO calibrates high-frequency responses inside the VSSD backbone, while OMP performs dynamic token generation and multi-prototype matching.}
  \label{fig:framework}
\end{figure}

In this section, we formulate the proposed CRISP (Calibration-aware Visual State Space Duality) framework for remote sensing semantic segmentation. First, in Section~\ref{sec:prelim}, we briefly review the State Space Duality (SSD) proposed in Mamba-2 and the mechanism causing feature information attenuation. Using its global-aggregation approximation as a starting point, we subsequently discuss the Duality Calibration Operator (DCO) in Section~\ref{sec:dco}, analyzing how it resolves spectral bias through a three-level hierarchical structure. Finally, in Section~\ref{sec:head}, we present the Orthogonal Multi-Prototype (OMP) head, demonstrating how it adapts to dispersed high-dimensional local characteristics via penalty regularizations.

\subsection{Mamba-2, VSSD, and Feature Smoothing Analysis}\label{sec:prelim}

Because state space models for vision are still relatively recent, we first give an intuitive picture before any formalism. A state space model (SSM) processes a sequence of tokens one position at a time, carrying a small running summary (a \emph{hidden state}) that is updated as it goes; reading this state out gives the output at each position. This makes the cost grow \emph{linearly} with the number of tokens, in contrast to self-attention, whose all-pairs comparison grows quadratically---which is why SSMs are attractive for the long token sequences produced by high-resolution remote sensing images. The Mamba family makes this update content-aware (so the model can decide what to keep or forget), and Mamba-2 shows that the whole process can be rewritten as a form of attention with a distance-based decay, which is both faster to compute and easier to reason about. The catch when moving to images is that pixels have no natural reading order, so visual variants such as VSSD scan the image along several directions and then \emph{fuse} the results into a single position-agnostic summary. As we make precise below, this fusion behaves like a large smoothing window: it captures global context well but averages away the fine high-frequency detail (edges, thin structures) that dense remote sensing segmentation depends on. The rest of this subsection formalizes this mechanism and pinpoints where the high-frequency loss originates, directly motivating our calibration design.

\textbf{Background: from selective SSMs to SSD.}
A state space model maps an input token sequence $x_1,\dots,x_L$ to outputs through a latent state $h_i$ that is updated recurrently, $h_i = \bar{A}h_{i-1} + \bar{B}x_i$, $y_i = C^\top h_i$, where $(\bar{A},\bar{B})$ are the discretized state-transition and input matrices and $C$ reads out the state. Selective SSMs (Mamba)~\cite{gu2024mamba} make $B$, $C$, and the discretization step \emph{input-dependent}, so that the model can route information selectively rather than with fixed dynamics. Mamba-2~\cite{dao2024transformers} introduces State Space Duality (SSD), showing that this recurrence is mathematically equivalent to a masked linear-attention form, which makes the connection to attention explicit and enables efficient parallel computation. We use this attention view below because it exposes \emph{how} the spatial mixing weights are formed, which is where over-smoothing originates.

In this dual form, the parameters $A,B,C \in \mathbb{R}^{L \times N}$ are produced from the input token features $X\in\mathbb{R}^{L\times D}$ by three separate linear projection layers, $B = X W_B$, $C = X W_C$, and the (input-dependent part of the) state decay $A = X W_A$, with learnable $W_A,W_B,W_C$; the values $V = X W_V \in \mathbb{R}^{L \times D}$ are likewise a linear projection of the input. Thus $B$/$C$ play the role of keys/queries and $A$ controls the distance decay. Given these, the sequential interactions in SSD can be uniformly expressed as a weighted sum in a prefix-scan kernel format. Specifically, the output feature at the $i$-th time step can be represented as:
\begin{equation}
  y_i = \sum_{j \le i} \kappa_{i,j} \langle c_i, b_j \rangle v_j
\end{equation}
where the kernel function $\kappa_{i,j}$ is implicitly determined by the state transition matrix and the discretization step size, reflecting a distance-dependent kernel that decays with sequence distance. The inner product $\langle c_i, b_j \rangle$ captures content-based interactions. The summation range $j \le i$ ensures strict causality, while the kernel $\kappa_{i,j}$ imparts a decay property that smoothly decreases as sequence distance increases.

However, 2D images inherently lack a unique and strict temporal orientation in spatial distribution. Therefore, architectures like Visual State Space Duality (VSSD) replace the single causal scan with a multi-directional sweeping and feature-fusion strategy. By relaxing the strictly causal lower-triangular mask while maintaining the $\mathcal{O}(L)$ computational bound, such models capture positional correlations through multi-directional scanning and subsequent aggregation. To facilitate a frequency-domain analysis of the impact of this aggregation, we abstract the multi-directional fusion operator into a global normalized aggregation approximation:
\begin{equation}\label{eq:vssd}
  \hat{y}_i \approx \frac{\sum_{j=1}^L w_{i,j} v'_j}{\sum_{j=1}^L w_{i,j}}, \quad w_{i,j} = \phi(\langle c_i, b_j \rangle),\ \phi(\cdot) \ge 0.
\end{equation}
In the equation above, $\phi(\cdot)$ is a non-negative monotonic mapping function (such as ReLU or exponential) and the resulting $w_{i,j}$ constitute the spatial fusion weights. The term $v'_j=\rho_j\odot v_j$ denotes the value after the per-token, input-dependent scaling that VSSD applies before aggregation, where $\rho_j$ is the gain (induced by discretization and decay) absorbed into the value path. We make this scaled value explicit here because the difference $v_j-v'_j$ is precisely the signal that our calibration branch recovers in Section~\ref{sec:dco}.

\textbf{Frequency-domain analysis of feature over-smoothing.}
We decompose each value into a slow-varying term and a local high-frequency perturbation, $v'_j = v^{\text{low}}_j + \epsilon_j$, where $\epsilon_j$ is approximately zero-mean within a local window. When the fusion weights $w_{i,j}$ vary smoothly over a large receptive field, the weighted summation in Equation~(\ref{eq:vssd}) averages out the oscillatory $\epsilon_j$, so the high-frequency energy in the output decays---manifesting as smoothed boundaries and details. Equivalently, with smooth weights the normalized aggregation $\alpha_{i,j}=w_{i,j}/\sum_j w_{i,j}$ approximates a large-window (in the uniform limit, $\alpha_{i,j}\!\approx\!1/L$, global) averaging operator, which is inherently low-pass and collapses the $L$ spatial tokens toward a low-rank summary. Crucially, once this aggregation has attenuated the high-frequency directions, position-dependent post-modulations alone cannot restore the lost amplitudes: an explicit high-frequency compensation branch is required. This provides a mechanistic explanation for the over-smoothing observed in visual Mamba baselines during dense remote sensing boundary delineation, and directly motivates DCO.

\subsection{Duality Calibration Operator}\label{sec:dco}

\begin{figure*}[!t]
  \centering
  \includegraphics[width=.92\textwidth,height=.78\textheight,keepaspectratio]{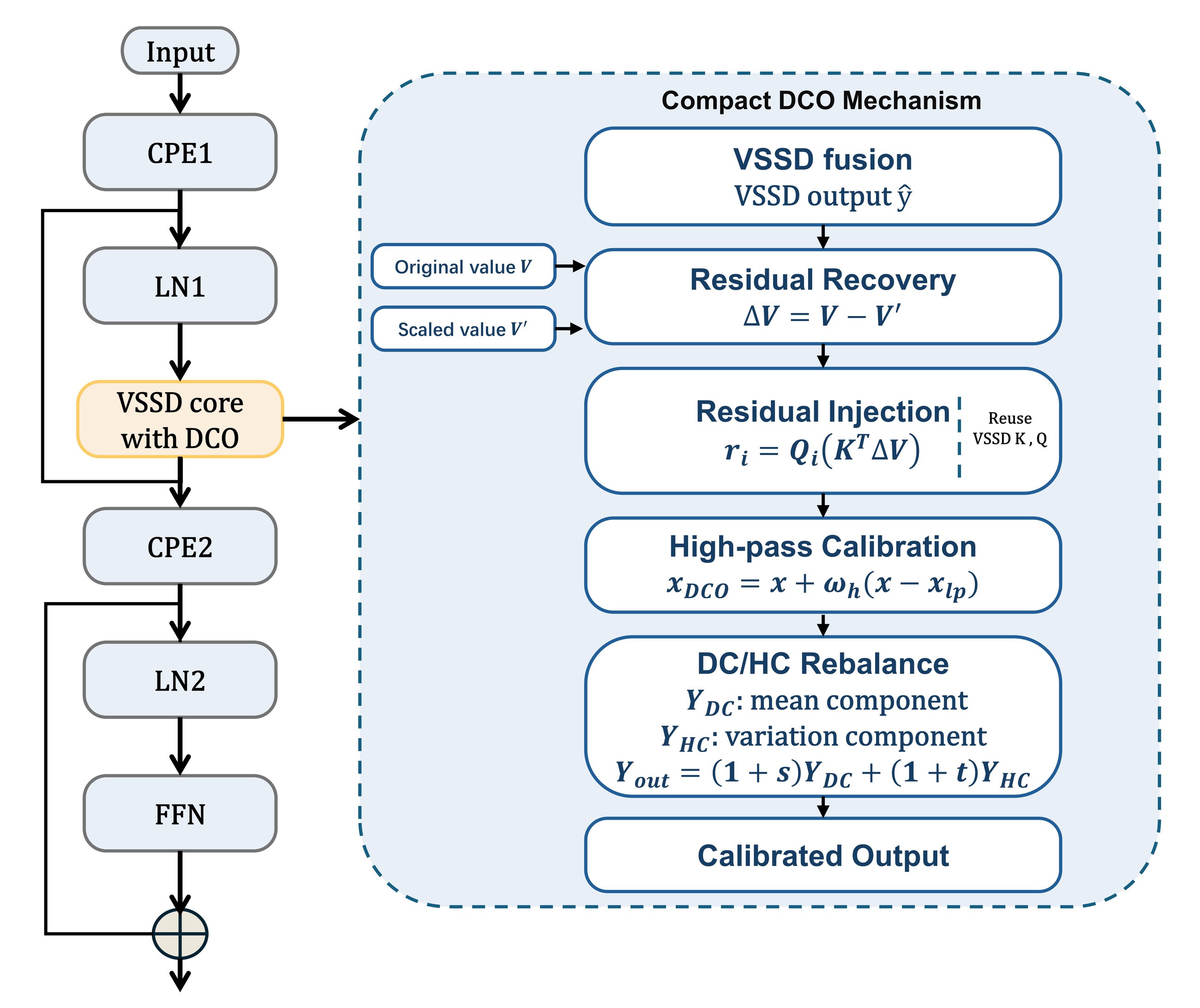}
  \caption{Compact illustration of DCO inside a VSSD block, including residual recovery, state-level injection, token-level high-pass calibration, and DC/HC rebalancing.}
  \label{fig:vssd_dco_arch}
\end{figure*}

DCO is inserted after VSSD's fusion operation, as illustrated in Fig.~\ref{fig:vssd_dco_arch}. It preserves VSSD's linear complexity by reusing the original aggregation weights, while compensating for the local contrast and boundary details weakened by global fusion. DCO has three near-identity calibration steps: \textbf{(i)} state-level residual injection for recovering contrast removed by value rescaling (Eqs.~(\ref{eq:dco_ri})--(\ref{eq:ri_linear})); \textbf{(ii)} token-level high-pass modulation for strengthening de-meaned responses (Eq.~(\ref{eq:high_pass})); and \textbf{(iii)} channel-level DC/HC rebalancing between low-frequency semantics and high-frequency details (Eq.~(\ref{eq:dco_out})).

\textbf{Residual extraction and linear-time injection.}
Following Section~\ref{sec:prelim}, the difference between the original and scaled values captures local detail weakened by VSSD fusion. We extract this residual variation as $\Delta v_j = v_j - v'_j$ and reuse the same kernelized key/query statistics already computed by VSSD. With $k_j=\psi_v(b_j)$, the residual statistics are accumulated as
\begin{equation}\label{eq:dco_ri}
\bm{S}_{\Delta}=\sum_{j=1}^{L}k_j\Delta v_j^\top .
\end{equation}
Since $\bm{S}_{\Delta}$ is shared across all query positions, it is computed once in linear time. The residual for token $i$ is then read out by the corresponding query feature $q_i=\psi_v(c_i)$:
\begin{equation}\label{eq:ri_linear}
r_i\approx q_i^\top \bm{S}_{\Delta},
\end{equation}
which uses the existing VSSD aggregation statistics and adds no quadratic interaction.
The residual is injected through a small-gain near-identity gate, $\tilde{y}_i=\hat{y}_i+(\varepsilon+\lambda_h)\odot r_i$, with $\lambda_h$ initialized to $0.01$. We use $\psi_v(\bm{x})=\mathrm{ELU}(\bm{x})+1$~\cite{katharopoulos2020transformers} and insert one DCO after the fusion operator in every VSSD block. Pseudocode, kernel choice, and complexity analysis are detailed in the supplementary material.

\textbf{Token-level frequency calibration.}
We further elevate fine-grained responses using a de-meaned high-pass modulation:
\begin{equation}\label{eq:high_pass}
y'_i = \tilde{y}_i + \omega_h \odot (\tilde{y}_i - \mu), \quad \text{with} \quad \mu = \frac{1}{L}\sum_{i=1}^{L} \tilde{y}_i,
\end{equation}
This is followed by a mean/variation decomposition $Y' = Y_{\text{dc}} + Y_{\text{hc}}$. Finally, DCO reallocates channel weights via
\begin{equation}\label{eq:dco_out}
Y_{\text{out}} = (1+s) \odot Y_{\text{dc}} + (1+t) \odot Y_{\text{hc}},
\end{equation}
where $\omega_h$, $s$, and $t$ are bounded and initialized small (to $0.01$). Thus, DCO starts near identity and then learns to rebalance low-frequency semantics and high-frequency structures during optimization.

\subsection{Orthogonal Multi-Prototype Decoder}\label{sec:head}

While DCO improves local separability, a single-prototype classifier may collapse rich intra-class modalities (e.g., ``same object, different spectra'') into a single direction. We propose an Orthogonal Multi-Prototype (OMP) decoder to preserve multi-modal structures in the logit space.

\textbf{Dynamic tokens and bidirectional interaction.}
As illustrated in the right inset of Fig.~\ref{fig:framework}, OMP generates content-adaptive class tokens in two steps: (1) a dual-path global pooling branch (max and average) extracts per-channel statistics from the feature map and fuses them through a shared MLP to produce content-adaptive query weights; (2) these weights are used to compose dynamic class tokens from a learned prototype embedding bank via a weighted Einsum. The resulting tokens are then refined with spatial features through stacked bidirectional cross-attention (details in supplementary).

\begin{figure}[t]
  \centering
  \includegraphics[width=\linewidth]{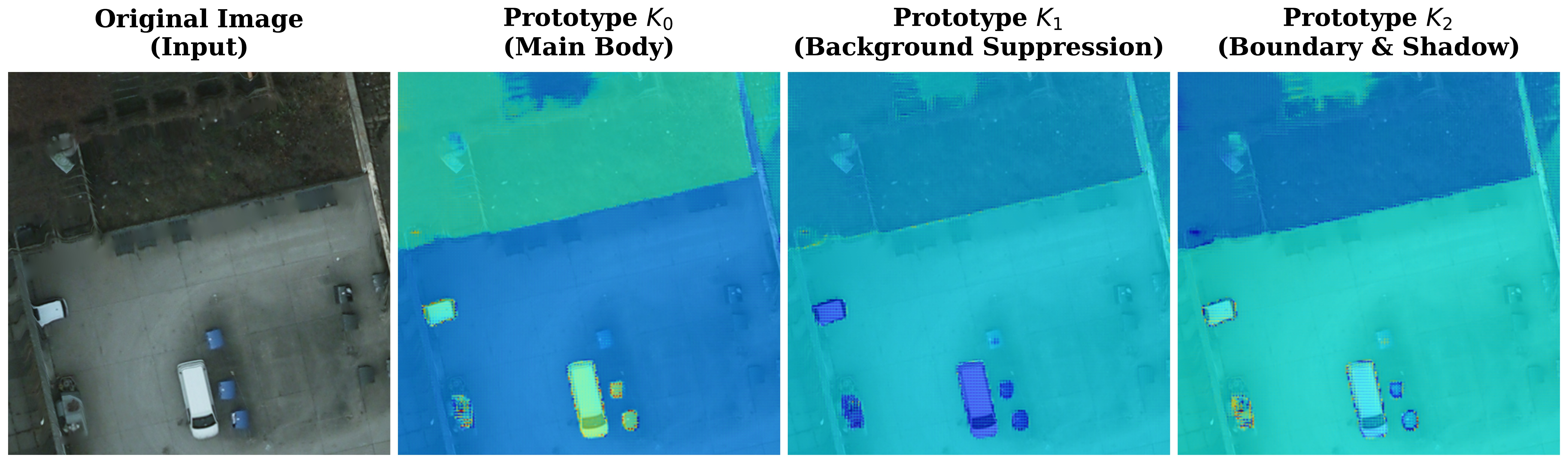}
  \caption{Sub-prototype spatial allocation on Potsdam ($K{=}3$). Different sub-prototypes capture distinct intra-class spectral modes.}
  \label{fig:proto_alloc}
\end{figure}

\textbf{Hyper-prototype matching.}
A hyper-network maps refined class tokens to $K$ prototypes per class $P\in\mathbb{R}^{B\times|\mathcal{C}|\times K\times d_h}$.
Here \emph{pixel descriptors} $\hat{U}\in\mathbb{R}^{(H'W')\times d_h}$ denote the per-location feature vectors of the decoder feature map (one $d_h$-dimensional vector per spatial position), after $L_2$ normalization; they are the entities matched against the prototypes.
As detailed in the left inset of Fig.~\ref{fig:framework}, given pixel descriptors $\hat{U}$ and normalized prototypes $\hat{P}$, the per-class confidence is computed by max-response matching across the $K$ sub-prototypes:
\begin{equation}\label{eq:omp_lse}
z_{c,i}=\max_{1\le k\le K}\langle \hat{U}_i,\hat{P}_{c,k}\rangle,
\end{equation}
so each pixel selects the closest sub-prototype of class $c$ while different sub-prototypes can specialize to different intra-class modes.
Figure~\ref{fig:proto_alloc} visualizes the spatial activation of each sub-prototype, confirming that different sub-prototypes specialize in distinct intra-class modes rather than collapsing to the same region.

\textbf{Dual prototype regularization.}
We impose intra-class orthogonality and inter-class angular margin constraints:
\begin{equation}\label{eq:loss_total}
\mathcal{L}_{\text{total}} = \mathcal{L}_{\text{seg}} + \lambda_{\text{orth}}\mathcal{L}_{\text{orth}} + \lambda_{\text{margin}}\mathcal{L}_{\text{margin}},
\end{equation}
where $\mathcal{L}_{\text{seg}}$ is the standard segmentation loss, $\mathcal{L}_{\text{orth}}$ penalizes redundant sub-prototypes, and $\mathcal{L}_{\text{margin}}$ enforces an angular margin $\delta$ between class centers $\hat{\bar{P}}_{b,c}$:
\begin{align}
\mathcal{L}_{\text{orth}} &= \frac{1}{B|\mathcal{C}|K(K-1)} \sum_{b,c} \sum_{k \neq l} \big| \langle \hat{P}_{b,c,k}, \hat{P}_{b,c,l} \rangle \big|, \label{eq:loss_orth} \\
\mathcal{L}_{\text{margin}} &= \frac{1}{B} \sum_{b} \sum_{c \neq c'} \mathrm{ReLU} \big( \langle \hat{\bar{P}}_{b,c}, \hat{\bar{P}}_{b,c'} \rangle - (1 - \delta) \big). \label{eq:loss_margin}
\end{align}
Sensitivity of mIoU to $K \in \{2,3,4,5\}$ on Potsdam is reported in Table S3; performance peaks at $K=3$ and remains stable for larger $K$, confirming that the gains come from multi-modal prototype coverage rather than increased model capacity. The margin $\delta$, loss weights $\lambda_{\text{orth}},\lambda_{\text{margin}}$, and calibration gains $\lambda_h,\omega_h,s,t$ are listed in the supplementary material.

\section{Experiments}\label{sec:exp}
\subsection{Experimental Setup}
We evaluate on ISPRS Potsdam and Vaihingen (5 classes, clutter ignored) and LoveDA (7 classes)~\cite{rottensteiner2012isprs,wang2021loveda}.
We report mIoU and mean F1-score (mF), and provide class-wise IoU for representative categories.
All models are re-trained under a unified MMSegmentation~\cite{contributors2021mmsegmentation} protocol with $512\times512$ random crops, AdamW (lr $6\times10^{-5}$, wd $0.05$)~\cite{loshchilov2019adamw}, linear warmup and Poly decay ($p=0.9$)~\cite{goyal2017accurate}.
We optimize the overall objective in Eq.~(\ref{eq:loss_total}). Additional training details (AMP, gradient clipping, and augmentation recipe) are deferred to the supplementary material.

\subsection{Comparison with State-of-the-Arts}

\begin{figure*}[t]
  \centering
  \includegraphics[width=0.92\textwidth]{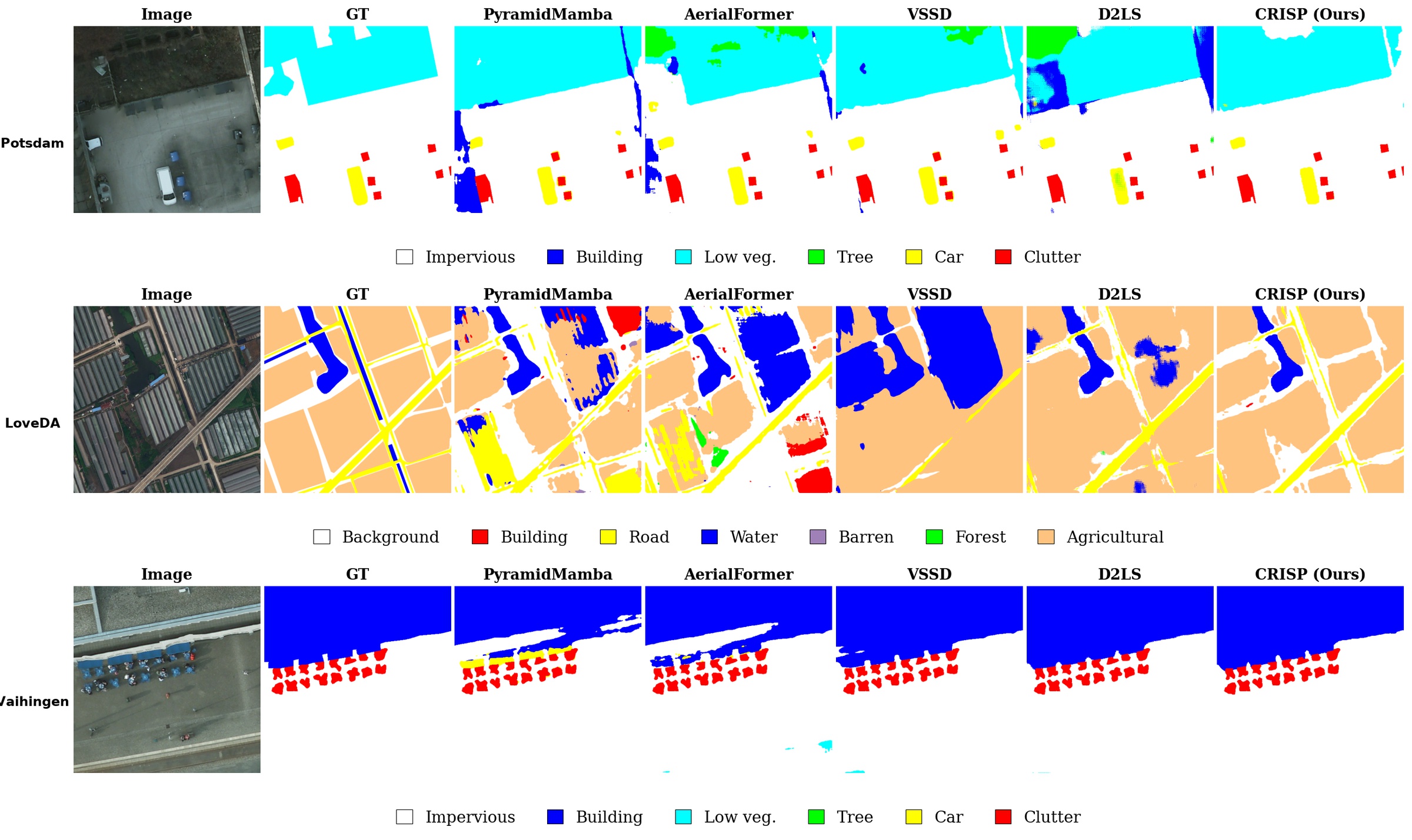}
  \caption{Qualitative comparison with VSSD on Potsdam, LoveDA, and Vaihingen. CRISP improves boundary delineation and local shape fidelity.}
  \label{fig:qualitative_sota}
\end{figure*}

\textbf{Quantitative and qualitative results.}
Table~\ref{tab:sota_loveda_fixed} reports LoveDA results, and Tables~\ref{tab:sota_potsdam_fixed} and~\ref{tab:sota_vaihingen_fixed} report Potsdam and Vaihingen respectively.
Across all benchmarks, CRISP achieves consistently strong mF and mIoU with a lightweight 32.32M parameter count.
Gains are most pronounced on categories with high intra-class variance and irregular contours (\eg, Low-veg and Tree on Potsdam/Vaihingen, Forest on LoveDA), which aligns directly with the design motivation of restoring attenuated high-frequency dynamics via DCO and preserving multi-modal structures via OMP.
Figure~\ref{fig:qualitative_sota} visualizes predictions on all three datasets: compared with the VSSD baseline, CRISP better preserves thin structures and object boundaries and reduces local shape distortion on small objects. Error map comparisons are in Fig.~S1.

\textbf{Efficiency.}
A full efficiency breakdown (parameters, FLOPs, peak memory, latency, and FPS) is provided in Table~S1.
Compared with the VSSD base model, CRISP reduces parameters and computation substantially while improving accuracy, indicating a favorable accuracy--efficiency trade-off rather than a purely capacity-driven gain.
The reduction is mainly attributed to replacing the heavy UPerHead with the lightweight MultiProtoDecoder, while DCO introduces negligible overhead ($<\!10^{-3}$M parameters and only $3.19$G FLOPs); a component-wise breakdown is given in Table~S2.

\begin{table*}[t]
  \centering
  \caption{Quantitative comparison on LoveDA. Class-wise IoU (\%) and mIoU are reported; best and second-best results are \textbf{bolded} and \underline{underlined}.}
  \label{tab:sota_loveda_fixed}
  \resizebox{\textwidth}{!}{
  \begin{tabular}{l|c|ccccccc|c}
    \hline
    \textbf{Method} & \textbf{Params (M)}  & \textbf{Bg} & \textbf{Bu} & \textbf{Rd} & \textbf{Wa} & \textbf{Ba} & \textbf{Fo} & \textbf{Ag} & \textbf{mIoU}\\
    \hline\hline
    ABCNet~\cite{li2021abcnet} & 13.6 & 49.78 & 46.56 & 48.08 & 63.83 & 20.03 & 38.33 & 47.04 & 44.81 \\
    VSSD~\cite{shi2025vssd} & 56.1 & 50.41 & 50.96 & 51.86 & 64.94 & 24.00 & 40.57 & 47.35 & 47.15 \\
    DCSwin~\cite{wang2022dcswin} & 45.6 & 49.70 & 56.95 & 47.92 & 66.14 & 27.53 & \underline{42.04} & 43.56 & 47.69 \\
    PyramidMamba~\cite{wang2025pyramidmamba} & 125.1 & 50.73 & 51.06 & 51.47 & \underline{67.81} & 26.78 & 40.08 & 49.17 & 48.16 \\
    BANet~\cite{wang2021transformer} & 12.7 & \underline{53.10} & 59.52 & 48.83 & 65.06 & 23.09 & 41.12 & 49.00 & 48.53 \\
    SFA-Net~\cite{hwang2024sfanet} & 10.7 & 52.58 & 58.73 & 53.53 & 66.84 & 27.31 & 39.16 & 48.85 & 49.57 \\
    UNetFormer~\cite{wang2022unetformer} & 11.7 & 51.29 & 58.89 & 53.76 & 66.92 & \underline{27.68} & 41.49 & 49.11 & 49.88 \\
    AerialFormer~\cite{hanyu2024aerialformer} & 43.2 & 52.09 & \underline{64.05} & \underline{55.21} & 64.31 & 27.40 & 41.78 & 48.07 & 50.41 \\
    \rowcolor{gray!10} \textbf{CRISP (Ours)} & 32.32 & \textbf{54.06} & \textbf{65.16} & 54.60 & 67.63 & 24.63 & \textbf{43.92} & \underline{50.90} & \underline{51.56} \\
    D2LS~\cite{Zou_2025_ICCV_D2LS} & 94.0 & 52.60 & 63.27 & \textbf{57.76} & \textbf{72.06} & \textbf{33.35} & 41.10 & \textbf{54.53} & \textbf{53.53} \\
    \hline
  \end{tabular}}
\end{table*}

\begin{table}[t]
  \centering
  \caption{Quantitative comparison on ISPRS Potsdam. Best/second-best results are \textbf{bolded}/\underline{underlined}.}
  \label{tab:sota_potsdam_fixed}
  \resizebox{\linewidth}{!}{
  \begin{tabular}{l|c|ccccc|cc}
    \hline
    \textbf{Method} & \textbf{Params (M)} & \textbf{Imp.} & \textbf{Build.} & \textbf{Low-veg.} & \textbf{Tree} & \textbf{Car} & \textbf{mFscore} & \textbf{mIoU} \\
    \hline\hline
    BANet & 12.7 & 92.31 & 94.42 & 86.27 & 84.93 & 94.95 & 90.58 & 83.04 \\
    PyramidMamba & 125.1 & 93.41 & 95.93 & 86.77 & 85.86 & 95.54 & 91.50 & 84.62 \\
    DCSwin & 45.6 & 93.57 & 96.13 & 87.51 & 87.03 & 96.12 & 92.07 & 85.56 \\
    VSSD & 56.1 & 93.78 & 96.27 & 87.13 & 87.20 & 95.97 & 92.07 & 85.57 \\
    SFA-Net & 10.7 & 94.06 & 96.63 & 87.96 & 87.28 & 96.46 & 92.48 & 86.28 \\
    ABCNet & 13.6 & 94.14 & 96.88 & 88.07 & 88.04 & 96.13 & 92.65 & 86.55 \\
    UNetFormer & 11.7 & 94.15 & 96.83 & 88.21 & 88.07 & 96.14 & 92.68 & 86.59 \\
    AerialFormer & 43.2 & 94.36 & 96.74 & 88.51 & 88.77 & 96.71 & 93.02 & 87.17 \\
    \rowcolor{gray!10} \textbf{CRISP (Ours)} & 32.32 & \underline{95.24} & \underline{97.78} & \underline{89.87} & \underline{89.47} & \underline{97.31} & \underline{93.93} & \underline{88.77} \\
    D2LS & 94.0 & \textbf{95.38} & \textbf{97.99} & \textbf{89.89} & \textbf{89.71} & \textbf{97.78} & \textbf{94.15} & \textbf{89.17} \\
    \hline
  \end{tabular}}
\end{table}

\begin{table}[t]
  \centering
  \caption{Quantitative comparison on ISPRS Vaihingen. Best/second-best results are \textbf{bolded}/\underline{underlined}.}
  \label{tab:sota_vaihingen_fixed}
  \resizebox{\linewidth}{!}{
  \begin{tabular}{l|c|ccccc|cc}
    \hline
    \textbf{Method} & \textbf{Params (M)} & \textbf{Imp.} & \textbf{Build.} & \textbf{Low-veg.} & \textbf{Tree} & \textbf{Car} & \textbf{mFscore} & \textbf{mIoU} \\
    \hline\hline
    SFA-Net & 10.7 & 91.76 & 94.35 & 82.42 & 88.43 & 75.73 & 86.54 & 76.87 \\
    UNetFormer & 11.7 & 92.23 & 94.65 & 83.75 & 89.04 & 79.16 & 87.77 & 78.64 \\
    DCSwin & 45.6 & 91.85 & 94.30 & 83.96 & 89.18 & 80.68 & 87.99 & 78.92 \\
    PyramidMamba & 125.1 & 92.38 & 94.99 & 83.49 & 88.88 & 81.41 & 88.23 & 79.32 \\
    BANet & 12.7 & 91.79 & 94.41 & 83.21 & 89.03 & 83.70 & 88.43 & 79.54 \\
    VSSD & 56.1 & 92.40 & 94.94 & 82.86 & 88.54 & 83.96 & 88.54 & 79.75 \\
    ABCNet & 13.6 & 92.82 & 95.44 & 83.29 & 89.06 & 84.86 & 89.09 & 80.64 \\
    AerialFormer & 43.2 & 92.80 & 95.75 & 83.88 & 89.00 & \underline{88.46} & 89.98 & 82.03 \\
    \rowcolor{gray!10} \textbf{CRISP (Ours)} & 32.32 & \underline{93.73} & \underline{96.62} & \underline{84.55} & \underline{89.46} & 88.37 & \underline{90.55} & \underline{83.00} \\
    D2LS & 94.0 & \textbf{93.74} & \textbf{96.95} & \textbf{84.73} & \textbf{89.71} & \textbf{90.79} & \textbf{91.18} & \textbf{84.06} \\
    \hline
  \end{tabular}}
\end{table}

\subsection{Ablation Studies}\label{sec:ablation}
We conduct ablations on the Potsdam validation set under single-scale inference.
Table~\ref{tab:ablation_dco} ablates the three DCO sub-components (residual injection, high-pass modulation, and DC/HC rebalancing), and Table~\ref{tab:ablation_components} evaluates end-to-end component synergy.
Activation-map visualizations are provided in Fig.~S2.

\begin{table}[t]
  \centering
  \caption{DCO sub-component ablation on Potsdam (UPerHead, single-scale).}
  \label{tab:ablation_dco}
  \footnotesize
  \setlength{\tabcolsep}{6pt}
  \begin{tabular}{l|ccc|c}
    \hline
    \textbf{Config.} & \textbf{Res.\ Inj.} & \textbf{Hi-Pass} & \textbf{DC/HC} & \textbf{mIoU} \\
    \hline
    Baseline              &            &            &            & 87.32 \\
    Res.\ Inj.\ only      & \checkmark &            &            & 87.65 \\
    Hi-Pass only          &            & \checkmark &            & 87.68 \\
    DC/HC only            &            &            & \checkmark & 87.64 \\
    Hi-Pass + DC/HC       &            & \checkmark & \checkmark & 87.83 \\
    \hline
    \textbf{Full DCO}     & \checkmark & \checkmark & \checkmark & \textbf{88.10} \\
    \hline
  \end{tabular}
\end{table}

\paragraph{DCO sub-components.}
Table~\ref{tab:ablation_dco} isolates the three calibration steps inside DCO. Each step, when activated alone, yields a consistent improvement over the 87.32 baseline (residual injection $+0.33$, high-pass modulation $+0.36$, DC/HC rebalancing $+0.32$), indicating that they address complementary facets of the over-smoothing problem rather than overlapping in effect. Residual injection restores the contrast removed by the per-token value rescaling, the high-pass term re-amplifies de-meaned boundary responses, and DC/HC rebalancing redistributes channel energy between low- and high-frequency components; none of them dominates in isolation. Combining the two frequency-domain steps (Hi-Pass\,+\,DC/HC) reaches 87.83, and adding residual injection on top recovers the remaining gap to the full DCO at 88.10. The fact that the full configuration exceeds the best two-way combination by a further $+0.27$ confirms that the state-level residual and the token-level frequency calibration are not redundant: the former supplies the high-frequency signal that the latter then re-weights. Because every gate is initialized near zero, DCO begins as an identity map and the improvements above are obtained without destabilizing the pretrained backbone.

\paragraph{Component synergy.}
\sloppy
Table~\ref{tab:ablation_components} shows that DCO alone improves the VSSD+UPerHead baseline from 87.32 to 88.10 mIoU, confirming that the gain is not merely due to decoder capacity. OMP alone underperforms the baseline (85.63), which is expected: without the high-frequency detail restored by DCO, the multiple sub-prototypes have little distinct intra-class structure to latch onto and the orthogonality constraint mainly disperses redundant directions. Once DCO is present, the picture reverses---OMP with both regularizers reaches the best 88.25, and the two regularizers contribute almost equally (orthogonality $88.06$, margin $88.02$ when applied singly), with their combination adding a further margin. This ordering (DCO $\rightarrow$ OMP $\rightarrow$ regularizers) indicates that multi-prototype matching is a beneficiary of, rather than a substitute for, the separability introduced by DCO: the calibrated features expose the intra-class modes, and the prototype regularizers keep those modes from collapsing back onto a single direction.
\fussy

\begin{table}[t]
  \centering
  \caption{End-to-end component ablation on Potsdam (VSSD backbone).}
  \label{tab:ablation_components}
  \footnotesize
  \setlength{\tabcolsep}{6pt}
  \renewcommand{\arraystretch}{1.1}
  \begin{tabular}{c|c|cc|c}
    \hline
    \textbf{Dec.} & \textbf{DCO} & \textbf{Orth} & \textbf{Margin} & \textbf{mIoU} \\
    \hline
    UPer &  &  &  & 87.32 \\
    UPer & \checkmark &  &  & 88.10 \\
    OMP &  & \checkmark & \checkmark & 85.63 \\
    OMP & \checkmark &  &  & 86.85 \\
    OMP & \checkmark & \checkmark &  & 88.06 \\
    OMP & \checkmark &  & \checkmark & 88.02 \\
    OMP & \checkmark & \checkmark & \checkmark & \textbf{88.25} \\
    \hline
  \end{tabular}
\end{table}

\FloatBarrier
\section{Conclusion}
\sloppy
This paper proposes CRISP to mitigate high-frequency attenuation and boundary blurring in visual state-space backbones. DCO treats the duality error of the state-space dual form as a measurable and correctable signal, restoring attenuated high-frequency responses through residual injection, local duality correction, and frequency-aware gating, while OMP preserves the recovered intra-class diversity with orthogonally regularized sub-prototypes. Experiments on remote-sensing semantic segmentation benchmarks demonstrate strong accuracy--efficiency trade-offs. Future work will explore spatially adaptive calibration conditioned on local frequency content.
\fussy

\section*{Acknowledgements}
\begin{sloppypar}
This work was supported by the Key Research Program of Hangzhou (No.~2025SZD2B02). It was also supported by the Major Science and Technology Program of Zhejiang Province (No.~2026LDC01007(JT)).
\end{sloppypar}

\bibliographystyle{splncs04}
\bibliography{main}
\end{document}

% Avoid hyperref PDF-string issue caused by "\\" in title
\title{\texorpdfstring{CRISP: Supplementary Material}{CRISP: Supplementary Material}}
\titlerunning{CRISP Supplementary Material}

% Author/affiliation identical to main.tex.
\author{Kangning Wang\inst{1,2} \and
Haopeng Zhang\inst{1,3,4}\thanks{Corresponding author.} \and
Zhiguo Jiang\inst{1}}
\authorrunning{K.~Wang et al.}
\institute{Tianmushan Laboratory, Beihang University, Hangzhou 311115, China \and
Hangzhou International Innovation Institute, Beihang University, Hangzhou 311115, China \and
School of Astronautics, Beihang University, Beijing 102206, China\\
\email{zhanghaopeng@buaa.edu.cn} \and
Key Laboratory of Spacecraft Design Optimization and Dynamic Simulation Technology,\\
Ministry of Education, Beijing 102206, China}
\maketitle

\section{Efficiency and Qualitative Evidence}
\label{sec:supp_eff_qual}

We report an efficiency breakdown on Potsdam and provide qualitative visualizations to complement the main paper.

\begin{table}[H]
  \centering
  \caption{Efficiency on Potsdam with $512\times512$ input.}
  \label{tab:efficiency_supp}
  \renewcommand{\arraystretch}{1.2}
  \resizebox{\linewidth}{!}{
  \begin{tabular}{lccccc}
    \toprule
    \textbf{Model} & \textbf{Param. (M)} & \textbf{FLOPs (G)} & \textbf{Mem. (G)} & \textbf{Lat. (ms)} & \textbf{FPS} \\
    \midrule
    SFA-Net       & \textbf{10.70} & \textbf{7.22}  & \underline{0.13} & 6.36  & 157.2 \\
    UNetFormer    & \underline{11.72} & \underline{11.76} & \textbf{0.10} & \underline{2.88}  & \underline{347.7} \\
    BANet         & 12.71             & 15.09          & 0.48          & 4.35  & 230.1 \\
    ABCNet        & 13.63             & 15.59          & 0.47          & \textbf{2.55}  & \textbf{392.1} \\
    \midrule
    AerialFormer  & 43.16             & 49.94          & 0.30          & 6.95  & 144.0 \\
    DCSwin        & 45.63             & 48.41          & 0.65          & 7.26  & 137.8 \\
    D2LS          & 94.03             & 93.57          & 1.49          & 10.79 & 92.7  \\
    PyramidMamba  & 125.11            & 118.73         & 31.03         & 12.16 & 82.2  \\
    \midrule
    VSSD (Base)   & 56.10             & 235.65         & 0.59          & 25.57 & 39.1  \\
    \rowcolor{gray!10} \textbf{CRISP (Ours)} & 32.32 & 64.80 & 0.43 & 22.03 & 45.4 \\
    \bottomrule
  \end{tabular}}
\end{table}

\begin{table}[H]
  \centering
  \caption{Component-wise parameter and FLOPs breakdown.}
  \label{tab:comp_breakdown}
  \small
  \setlength{\tabcolsep}{6pt}
  \renewcommand{\arraystretch}{1.15}
  \begin{tabular}{lcccc}
    \hline
    \textbf{Method / Metric} & \textbf{Encoder} & \textbf{DCO} & \textbf{Decoder} & \textbf{Total} \\
    \hline
    VSSD baseline FLOPs (G) & 25.07 & -- & 210.58 & 235.65 \\
    \hline
    CRISP Params (M)        & 26.20 & $\!<\!10^{-3}$ & 6.13  & 32.32 \\
    CRISP FLOPs (G)         & 45.83 & 3.19          & 15.78 & 64.80 \\
    \hline
  \end{tabular}
\end{table}

\noindent
Tables~\ref{tab:efficiency_supp} and~\ref{tab:comp_breakdown} show that CRISP mainly reduces cost by replacing the heavy UPerHead-style decoder with a lightweight multi-prototype decoder, while DCO itself contributes negligible parameters and modest FLOPs. Compared with the VSSD baseline, CRISP cuts the total FLOPs from $235.65$G to $64.80$G and reduces parameters from $56.10$M to $32.32$M, while also improving FPS under the same evaluation setting.

\paragraph{On the latency of state-space operators.}
As shown in Table~\ref{tab:efficiency_supp}, CRISP is not latency-optimal: its current latency disadvantage stems largely from the still-immature CUDA kernels for state-space operators, whereas convolution/attention kernels have benefited from years of engineering (\eg, TensorRT). This gap is an implementation artifact rather than a property of the model, and is expected to narrow as SSM kernels mature. The practical payoff of CRISP lies in deployment settings where \emph{static and runtime memory}, not raw latency, is the hard constraint---such as on-board UAV/satellite processors with limited on-chip SRAM, where a $1.49$G footprint can simply fail to fit while a $0.43$G one runs. In such regimes, parameter and memory efficiency at real-time speed is the decisive factor.

\begin{figure}[t]
  \centering
  \includegraphics[width=\linewidth]{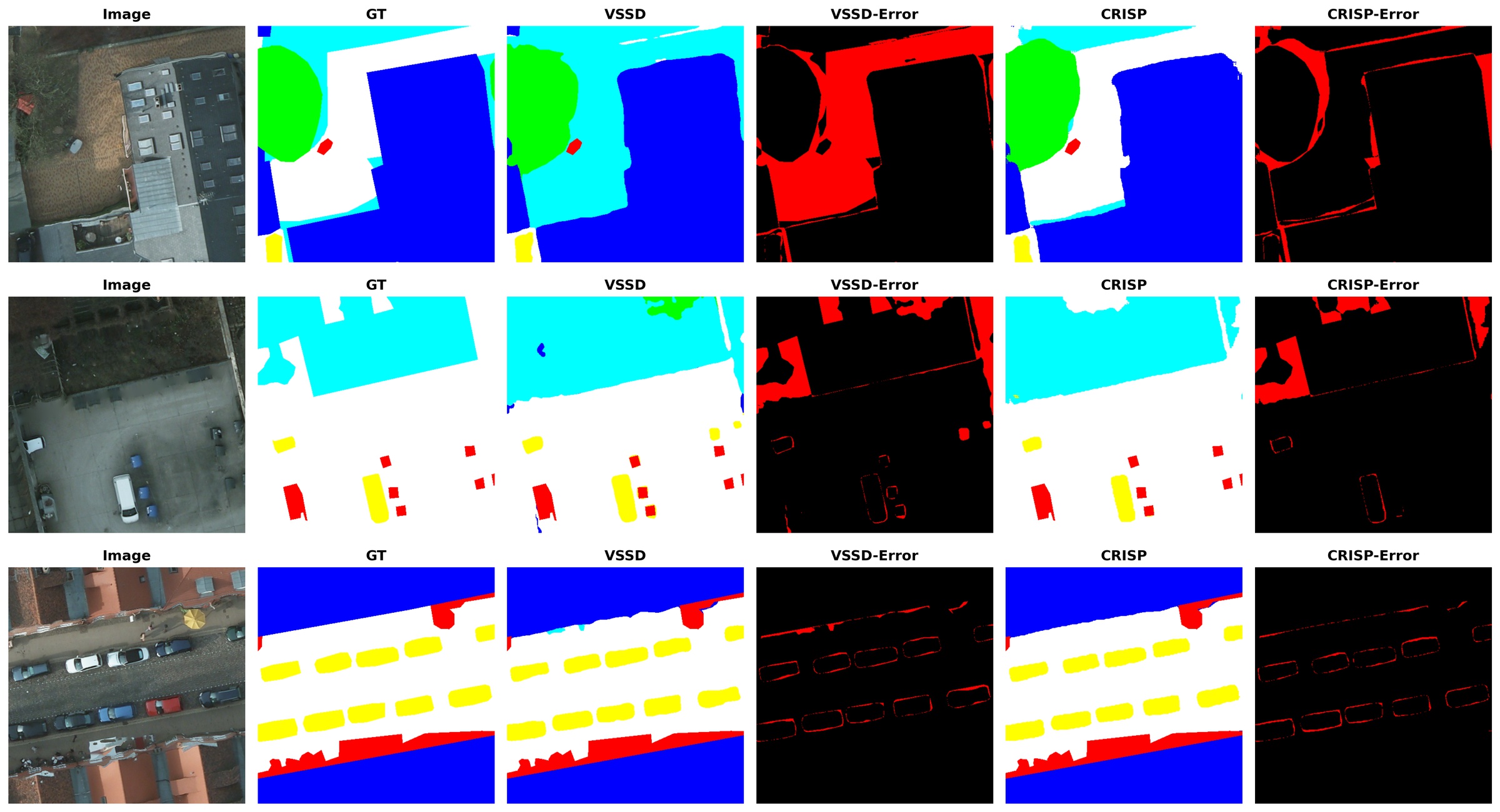}
  \caption{Error maps comparing VSSD and CRISP. Red pixels indicate prediction errors.}
  \label{fig:error_map}
\end{figure}

\begin{figure}[t]
  \centering
  \includegraphics[width=.82\linewidth]{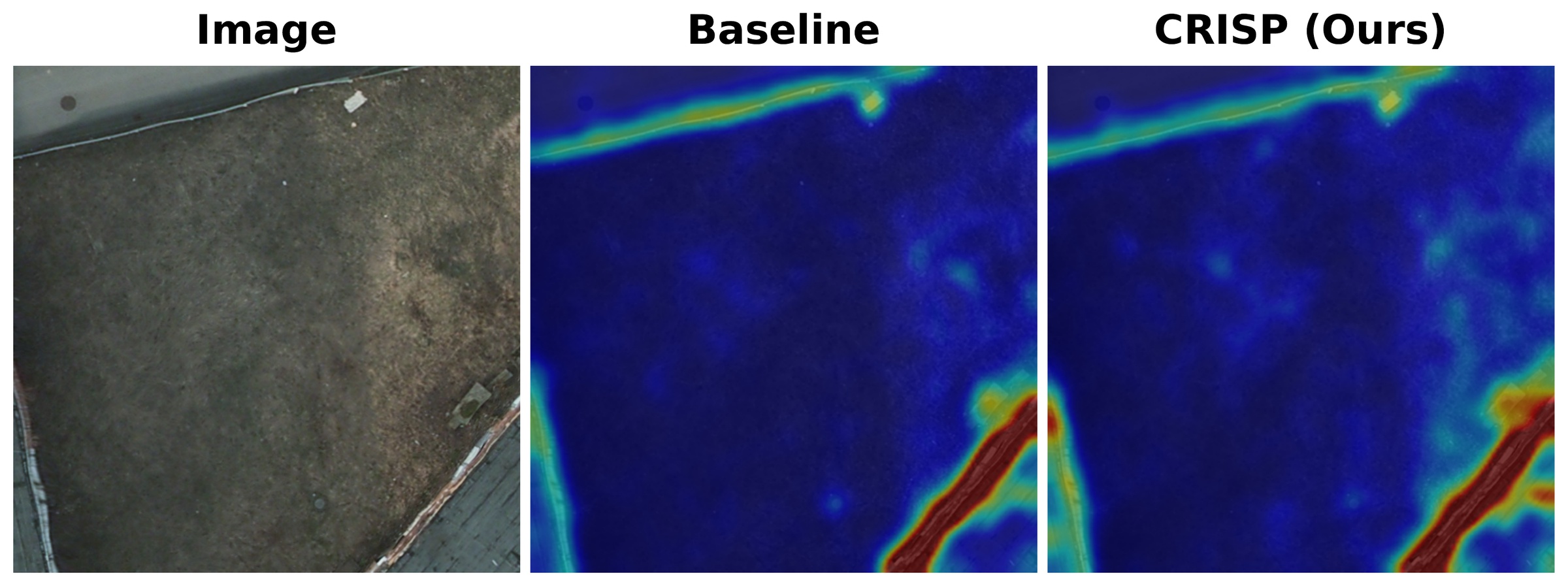}
  \caption{Feature activation maps before and after DCO.}
  \label{fig:activation_map}
\end{figure}

\noindent
Figures~\ref{fig:error_map} and~\ref{fig:activation_map} provide visual evidence for the same trend. The VSSD baseline tends to produce fragmented errors around thin objects, roof boundaries, and small structures, whereas CRISP reduces these localized errors. The activation maps further show that DCO strengthens boundary-aware responses instead of only changing the final classifier output.

% ---------------------------------------------------------------
\section{Effect of Prototype Number $K$}
\label{sec:supp_k}

Table~\ref{tab:supp_k} reports single-scale validation mIoU and mFscore on Potsdam for
$K \in \{2, 3, 4, 5\}$ with all other hyper-parameters fixed.
Performance peaks at $K{=}3$ and plateaus for $K{\ge}4$, indicating that three sub-prototypes
per class are sufficient to cover dominant intra-class modes without overfitting.
(Sub-prototype spatial allocation visualizations are provided in the main paper.)

\begin{table}[H]
  \centering
  \caption{Effect of prototype number $K$ on Potsdam.}
  \label{tab:supp_k}
  \renewcommand{\arraystretch}{1.2}
  \begin{tabular}{ccc}
    \toprule
    \textbf{$K$} & \textbf{mIoU (\%)} & \textbf{mFscore (\%)} \\
    \midrule
    2 & 88.10 & 93.56 \\
    3 & \textbf{88.25} & \textbf{93.64} \\
    4 & 88.17 & 93.59 \\
    5 & 88.19 & 93.60 \\
    \bottomrule
  \end{tabular}
\end{table}

% ---------------------------------------------------------------
\section{Extended Boundary and Cross-Dataset Analysis}
\label{sec:supp_boundary}

\paragraph{Boundary-band diagnostic.}
Complementing the main paper, Table~\ref{tab:supp_boundary_band} reports the boundary-band diagnostic on the Potsdam validation set. We dilate ground-truth semantic boundaries into bands and evaluate predictions only within these regions. Removing either DCO or the prototype regularizers consistently degrades boundary-band mIoU, boundary-band accuracy, and mBF, while the full model is best on all three metrics. This supports that both components contribute to boundary preservation rather than acting as unrelated add-ons.

\begin{table}[H]
  \centering
  \caption{Boundary-band diagnostic on Potsdam.}
  \label{tab:supp_boundary_band}
  \renewcommand{\arraystretch}{1.2}
  \begin{tabular}{lccc}
    \toprule
    \textbf{Model} & \textbf{Band mIoU} & \textbf{Band Acc} & \textbf{mBF} \\
    \midrule
    w/o DCO            & 0.6830 & 0.7833 & 0.8698 \\
    w/o proto.\ regs   & 0.6848 & 0.7840 & 0.8232 \\
    \rowcolor{gray!10} \textbf{Full CRISP} & \textbf{0.7179} & \textbf{0.8104} & \textbf{0.8953} \\
    \bottomrule
  \end{tabular}
\end{table}

\paragraph{Cross-dataset boundary quality.}
Table~\ref{tab:supp_mbf_cross} reports the class-aware mean Boundary F-score (mBF) on all three datasets referenced in the main paper. CRISP consistently outperforms the VSSD baseline on every dataset, confirming that the boundary-delineation gains generalize beyond a single benchmark.

\begin{table}[H]
  \centering
  \caption{Cross-dataset class-aware mBF (CRISP vs.\ VSSD baseline).}
  \label{tab:supp_mbf_cross}
  \small
  \setlength{\tabcolsep}{8pt}
  \begin{tabular}{lccc}
    \toprule
    \textbf{Dataset} & \textbf{VSSD} & \textbf{CRISP} & \textbf{Gain} \\
    \midrule
    Potsdam   & 0.6669 & \textbf{0.7038} & \textbf{+0.0369} \\
    Vaihingen & 0.6730 & \textbf{0.6935} & \textbf{+0.0205} \\
    LoveDA    & 0.1999 & \textbf{0.2239} & \textbf{+0.0240} \\
    \bottomrule
  \end{tabular}
\end{table}

% ---------------------------------------------------------------
\section{DCO Implementation Details}
\label{sec:supp_dco}

\begin{figure}[H]
  \centering
  \includegraphics[width=.78\linewidth]{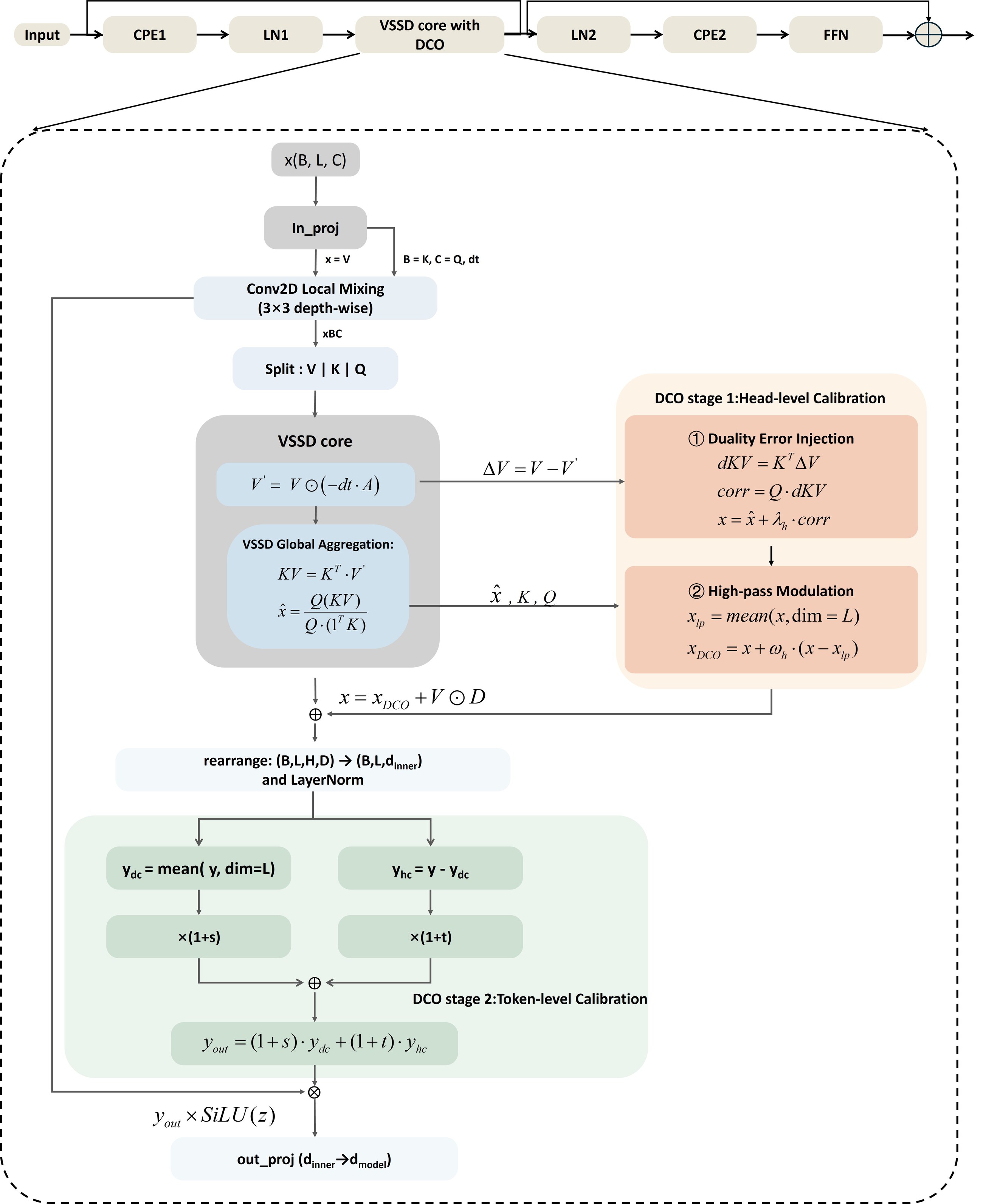}
  \caption{Detailed VSSD core with DCO, complementing the compact main-paper diagram.}
  \label{fig:supp_dco_detail}
\end{figure}

This section expands the compact DCO diagram in the main paper by pairing the module view in Fig.~\ref{fig:supp_dco_detail} with the step-by-step pseudocode in Algorithm~\ref{alg:dco}. The figure shows where residual recovery, high-pass calibration, and DC/HC rebalancing are inserted around the VSSD core, while the pseudocode specifies how these operations are computed with the reused VSSD aggregation statistics.

\FloatBarrier

\subsection{Kernel and Numerical Stability}

\noindent\textbf{Insertion.} Unless otherwise stated, we insert one DCO after the VSSD fusion operator in every VSSD block (all stages).

In the main text, the linear-time residual injection reuses the VSSD $Q,K$ tensors and requires a decomposable kernel $\psi_v(\cdot)$ that keeps inner products non-negative.
We use the ELU+1 kernel~\cite{katharopoulos2020transformers}:
\begin{equation}
  \psi_v(\bm{x}) = \mathrm{ELU}(\bm{x}) + 1 =
  \begin{cases}
    \bm{x} + 1 & \bm{x} \ge 0 \\
    e^{\bm{x}} & \bm{x} < 0
  \end{cases}
\end{equation}
The base VSSD fusion still clamps its row-normalization denominator with $\epsilon_d{=}10^{-6}$ (float32) or $10^{-4}$ (bfloat16/float16), while the DCO residual injection itself uses the unnormalized read-out $Q_i^\top\bm{S}_{\Delta}$.

\subsection{Pseudocode and Complexity}

The pseudocode in Algorithm~\ref{alg:dco} follows the data flow in Fig.~\ref{fig:supp_dco_detail}: it first recovers the duality residual from the original and discretized values, injects this residual through the kernelized aggregation, then applies high-pass calibration and DC/HC rebalancing before returning the calibrated output.
The accumulator $\bm{S}_{\Delta}$ is computed in a single pass, adding
$\mathcal{O}(L\cdot d_{\text{state}}\cdot D)$ with no asymptotic overhead beyond the existing linear attention pass.
The extra parameters per block are $\mathcal{O}(H{+}D)$ (gates), negligible compared to $\mathcal{O}(D^2)$ projections.
In other words, DCO reuses the same kernelized aggregation statistics already available in the VSSD computation and only adds lightweight residual and channel-wise gates.

\begin{algorithm}[H]
\caption{DCO Forward Pass \quad (per VSSD block)}
\label{alg:dco}
\begin{algorithmic}[1]
\Require $V \in \mathbb{R}^{B \times H \times L \times D}$,\;
         $K = \psi_v(B)$,\; $Q = \psi_v(C) \in \mathbb{R}^{B \times 1 \times L \times N}$,\;
         $\hat{y} \in \mathbb{R}^{B \times H \times L \times D}$
\Statex \textbf{Param:} $\lambda_h, \omega_h \in \mathbb{R}^{H}$;\quad $s, t \in \mathbb{R}^{D}$ \;(small-gain init)
\State $V' \leftarrow V \cdot \Delta A$ \Comment{existing discretised value}
\State $\Delta V \leftarrow V - V'$ \Comment{duality error}
\State $\bm{S}_\Delta \leftarrow \sum_{j} K_j\,\Delta V_j^\top \in \mathbb{R}^{N \times D}$
\State $r_i \leftarrow Q_i^\top \bm{S}_\Delta$ \Comment{residual injection}
\State $\tilde{y}_i \leftarrow \hat{y}_i + (\varepsilon+\lambda_h) \odot r_i$,\;\; $\varepsilon{=}10^{-4}$ \Comment{near-identity init}
\State $\mu \leftarrow \tfrac{1}{L}\sum_i \tilde{y}_i$;\quad
       $\tilde{y}_i \leftarrow \tilde{y}_i + \omega_h \odot (\tilde{y}_i - \mu)$ \Comment{high-pass modulation}
\State $Y' \leftarrow \{\tilde{y}_i\}_{i=1}^{L}$
\State $Y_\text{dc} \leftarrow \mu \cdot \mathbf{1}^\top$;\quad $Y_\text{hc} \leftarrow Y' - Y_\text{dc}$
\State \Return $(1{+}s) \odot Y_\text{dc} + (1{+}t) \odot Y_\text{hc}$ \Comment{DC/HC rebalance}
\end{algorithmic}
\end{algorithm}

\bibliographystyle{splncs04}
\bibliography{main}